\documentclass[letterpaper, 10 pt, conference]{ieeeconf}  % Comment this line out if you need a4paper

\IEEEoverridecommandlockouts                              % This command is only needed if 
\usepackage{amssymb,amsmath}
\usepackage{booktabs}
\usepackage{caption}
\usepackage{subcaption}
\usepackage{graphicx}
\usepackage{graphics}

\usepackage{multicol,multirow}
\usepackage{verbatimbox}

\usepackage{hyperref}
\hypersetup{bookmarksopen,bookmarksnumbered,
pdfpagemode=UseOutlines,
colorlinks=true,
linkcolor=blue,
anchorcolor=blue,
citecolor=blue,
filecolor=blue,
menucolor=blue,
urlcolor=blue
}

\title{\LARGE \bf
Towards Rich, Portable, and Large-Scale Pedestrian Data Collection*
}

\author{Allan Wang$^{1}$, Abhijat Biswas$^{1}$, Henny Admoni$^{1}$ and Aaron Steinfeld$^{1}$% <-this % stops a space
\thanks{*This work was supported by grant (IIS-1734361) and (IIS-1900821) from the National Science Foundation}% <-this % stops a space
\thanks{$^{1}$The authors are with the Robotics Institute, Carnegie Mellon University,
        5000 Forbes Avenue, Pittsburgh, PA, USA 
        {\tt\small \{allanwan, abhijatb\}@andrew.cmu.edu, 
        \{henny, steinfeld\}@cmu.edu}}%
}

\begin{document}

\maketitle
\thispagestyle{empty}
\pagestyle{empty}

%%%%%%%%%%%%%%%%%%%%%%%%%%%%%%%%%%%%%%%%%%%%%%%%%%%%%%%%%%%%%%%%%%%%%%%%%%%%%%%%
\begin{abstract}

Recently, pedestrian behavior research has shifted towards machine learning based methods and converged on the topic of modeling pedestrian interactions. For this, a large-scale dataset that contains rich information is needed. We propose a data collection system that is portable, which facilitates accessible large-scale data collection in diverse environments. We also couple the system with a semi-autonomous labeling pipeline for fast trajectory label production. We further introduce the first batch of dataset from the ongoing data collection effort -- the TBD pedestrian dataset. Compared with existing pedestrian datasets, our dataset contains three components: human verified labels grounded in the metric space, a combination of top-down and perspective views, and naturalistic human behavior in the presence of a socially appropriate ``robot".

\end{abstract}

%%%%%%%%%%%%%%%%%%%%%%%%%%%%%%%%%%%%%%%%%%%%%%%%%%%%%%%%%%%%%%%%%%%%%%%%%%%%%%%%
\section{Introduction}

Pedestrian datasets are essential tools for designing socially appropriate robot behaviors, recognizing and predicting human actions, and studying pedestrian behavior. A generally accepted assumption for these datasets is that real-world pedestrians are experts in analyzing and navigating human crowds because they are proficient at behaving in accordance to social interaction norms. 
%Behavioral or practical research related to pedestrian motion likely involves constructing a model that captures these social interactions and movements. In general, existing datasets have been collected to support specific research questions, leading to inadvertent limitations on utility towards certain research questions. This paper describes our efforts to collect and create a dataset that supports a larger array of research questions.

\begin{figure}[h]
      \centering
      \includegraphics[scale=0.36]{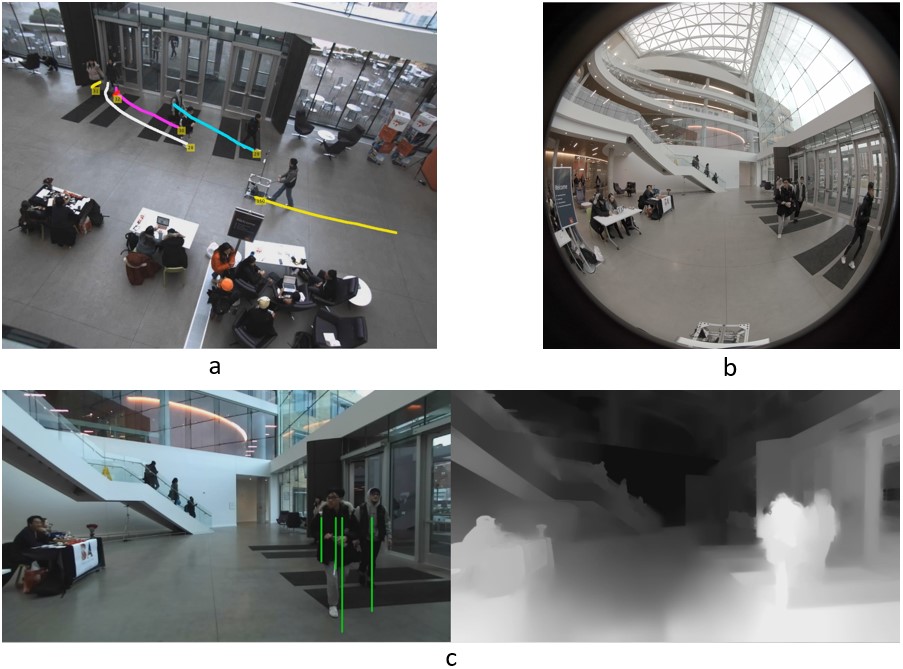}
      \caption{This set of images represent the same moment recorded from multiple sensors: a) Top-down view image taken by a static camera with ground truth pedestrian trajectory labels shown. b) Perspective-view image from a 360 camera that captures high definition videos of nearby pedestrians. c) Perspective-view RBG and depth images from a stereo camera mounted on a cart that is used to imitate onboard robot sensors. Green vertical bars represent the projected labels. Note that two pedestrians at the back are partially and completely occluded from the stereo camera.}
      \label{fig:intro}
      \vspace{-1em}
  \end{figure}

Researchers may use these data to predict future pedestrian motions, including forecasting their trajectories \cite{Alahi1, Gupta1, ivanovic-trajectron}, and/or navigation goals \cite{kitani-2012, liang2020garden}. In social navigation, datasets can also be used to model interactions. For example, a key problem researchers have tried to address is the \textit{freezing robot problem} \cite{Trautman1}, in which the robot becomes stuck in dense, crowded situations while trying to be deferential to human movements for safety or end user acceptance reasons. Researchers have attributed this problem to robot's inability to model interactions \cite{sun2021move}. In other words, most current navigation algorithms do not consider pedestrian reactions and assume a non-cooperative environment. Some works \cite{nishimura2020risk} have used datasets to show that modeling the anticipation of human reactions to the robot's actions enables the robot to deliver a better performance.

%However, interactions are diverse and can be rare occurrences in human crowds. Although robotic systems typically have access to each pedestrian's basic properties (e.g., position and velocity), inter-pedestrian interactions are less frequent because interactions require the presence of two or more pedestrians that usually need to be in close proximity of each other. While data documenting interactions is more limited, some work has made progress on this front. For example, Sch\"oller et al. \cite{constant-velocity-model} has shown that a linear acceleration based method can perform comparably with deep learning based models in pedestrian trajectory prediction settings. This implies that pedestrians mostly walk in linear fashion, a default behavior when not interacting with other pedestrians. Additionally, pedestrian interactions can be very diverse, especially in certain contexts. Some categories of interactions that researchers have devised include collision avoidance, grouping \cite{wang-split-merge}, and leader-follower \cite{kothari2021human}. The details of these types of interactions can further be diversified by the environment (e.g. an open plaza or a narrow corridor). Mavrogiannis et al. \cite{mavrogiannis_etal2021-core-challenges} provides more details on interaction types. 

In order to better capture and model interactions to improve the performance of various pedestrian-related algorithms, considerably more data is needed across a variety of environments. To this end, we have constructed a data collection system that can achieve these two requirements: large quantity and environment diversity. First, we ensure that our equipment is completely portable and easy to set up. This allows collecting data in a variety of locations with limited lead time. Second, we address the challenge of labeling large quantities of data using a semi-autonomous labelling pipeline. We employ a state-of-the-art deep learning based \cite{zhang2021bytetrack} tracking module combined with various post-processing procedures to automatically produce high quality ground truth pedestrian trajectories in metric space. 

We hope our dataset approach offers various improvements and aims to accommodate a wide variety of pedestrian behavior research. Specifically, we include three important characteristics: (1) ground truth labeling in metric space, (2) perspective views from a moving agent, and (3) natural human motion. To the best of our knowledge, publicly available datasets only have at most two of these characteristics, but not all three. We demonstrate our system through a dataset collected in a large indoor space: the TBD pedestrian dataset\footnote{\href{https://tbd.ri.cmu.edu/tbd-social-navigation-datasets}{https://tbd.ri.cmu.edu/tbd-social-navigation-datasets}}. Our dataset contains scenes with a variety of crowd densities and pedestrian interactions. This dataset can be used to complement existing datasets by injecting a new data environment and more pedestrian behavior distribution into existing dataset mixtures, such as \cite{kothari2021human}. This is an ongoing effort and we have only released the first dataset batch.

\section{System Description}\label{sec:system}

%In this work, we introduce a data collection system that is portable and easy to setup that will allow easy collection of large quantities of data. The data collection setup also contains a cart that provides data on naturalistic pedestrian reactions to the robot from a typical perspective view. 

\subsection{Hardware Setup}\label{sec:hardware}

\begin{figure}[tbhp]
    \centering
    \includegraphics[scale=0.3]{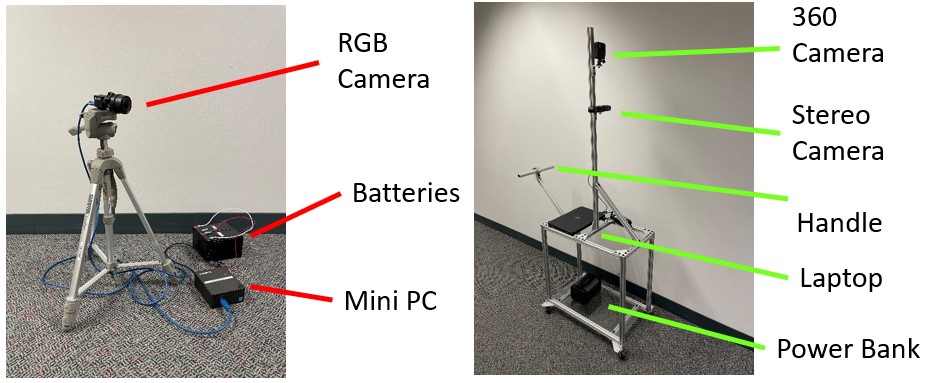}
    \caption{Sensor setup used to collect the TBD pedestrian dataset. (left) one of three nodes used to used to capture top-down RGB views. Each node is self contained with an external battery and communicates wirelessly with other nodes.
    (right) cart used to capture sensor views from the mobile robot perspective during data collection. The cart is powered by an onboard power bank and laptop.}
    \label{fig:camera}
\end{figure}

\begin{figure}[]
    \centering
    \includegraphics[scale=0.25]{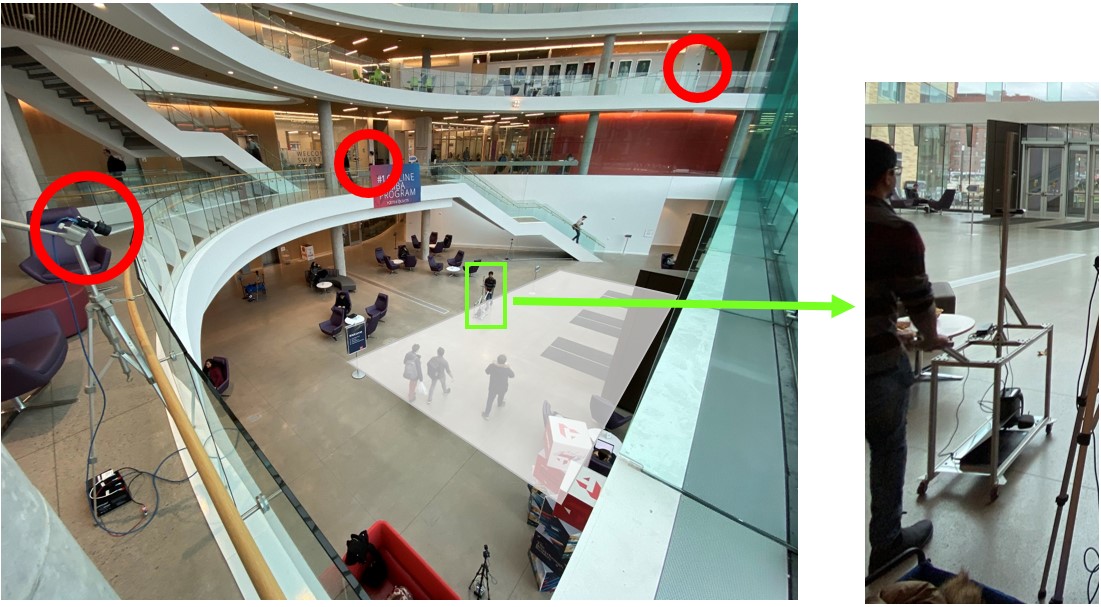}
    \caption{Hardware setup for the TBD pedestrian dataset. Red circles indicate positions of RGB cameras. Green box shows our mobile cart with a 360 camera and stereo camera which imitate a mobile robot sensor suite. The cart is manually pushed by a researcher during recording. The white area is where trajectory labels are collected.}
    \label{fig:hdc_setup}
    \vspace{-1em}
\end{figure}

As shown in Figure~\ref{fig:hdc_setup}, we positioned three FLIR Blackfly RGB cameras (Figure~\ref{fig:camera}) surrounding the scene on the upper floors overlooking the ground level at roughly 90 degrees apart from each other. The RGB cameras are connected to portable computers powered by lead-acid batteries. We also positioned three more units on the ground floor, but did not use them for pedestrian labeling. 
% Compared to a single overhead camera, multiple cameras ensure better pedestrian labeling accuracy. This is achieved by labeling the pedestrians from cameras that have the highest image resolution of the pedestrians (i.e., closest to pedestrians).

In addition to the RGB cameras, we pushed a cart through the scene (Figure~\ref{fig:camera}) equipped with a ZED stereo camera to collect both perspective RGB views and depth information of the scene. A GoPro Fusion 360 camera for capturing high definition 360 videos of nearby pedestrians was mounted above the ZED. Data from the on-board cameras are useful in capturing pedestrian pose data and facial expressions. The ZED camera was powered by a laptop with a power bank. Our entire data collection hardware system is portable and does not require power outlets, thereby allowing data collection outdoors or in areas where wall power is inaccessible.
%Besides the cameras, the cart is also fitted with a handle so that one of our researchers can push the cart through the environment.

During each data collection session, we pushed the cart from one end of the scene to another end, while avoiding pedestrians and obstacles along the way in a natural motion similar to a human pushing a delivery cart. The purpose of this cart was to represent a mobile robot traversing through the human environment. However, unlike other datasets such as \cite{lcas} or \cite{jrdb} that use a Wizard-of-Oz controlled robot, we used a manually pushed cart. This provided better trajectory control, increased safety, and reduced the novelty effect from pedestrians, as curious pedestrians may intentionally block robots or display other unnatural movements \cite{brvsvcic2015escaping}.

The first batch of our data collection occurred on the ground level in a large indoor atrium area (Figure~\ref{fig:hdc_setup}). Half of the atrium area had fixed entry/exit points that led to corridors, elevators, stairs, and doors to the outside. The other half of the atrium was adjacent to another large open area and was unstructured with no fixed entry/exit points. We collected data around lunch and dinner times to ensure higher crowd densities. More data will be collected in the future in locations such as transit stations.

\subsection{Post-processing and Labeling}

\begin{figure}[thpb]
      \centering
      \includegraphics[scale=0.25]{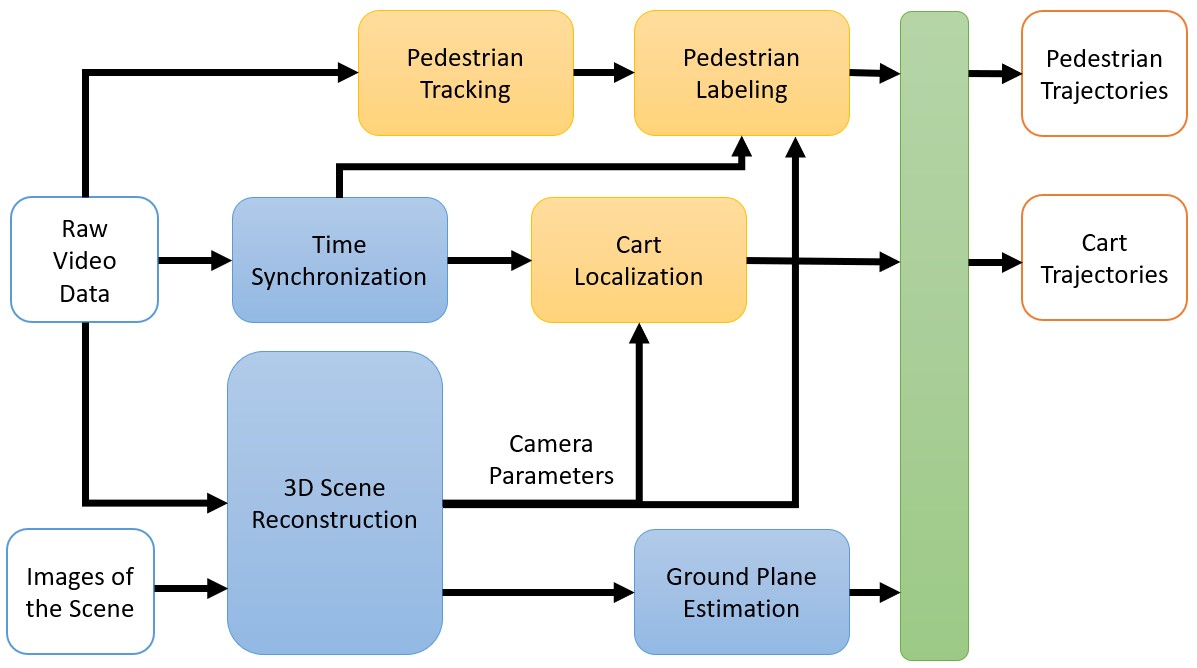}
      \caption{Flowchart for our post-processing pipeline. Blue blocks are preparation procedures and orange blocks are labeling procedures. The green block transforms all trajectory labels onto the ground plane $z=0$.}
      \label{fig:system-flowchart}
   \end{figure}

A summary of our post processing pipeline is summarized in Figure \ref{fig:system-flowchart}. We expand on select nodes to explain the post-processing procedures in greater detail.

\label{sec:postprocessing}

\subsubsection{Time synchronization}\label{sec:calibration}
To ensure time synchronization across the captured videos, we employed Precision Time Protocol over a wireless network to synchronize each of the the computers powering the cameras, which allows for sub-microsecond synchronization. For redundancy, we held an LED light at a location inside the field of view of all the cameras and switched it on and off at the beginning of each recording session. We then checked for the LED light signal during the post-processing stage to synchronize the starting frame of all the captured videos for each recording session.

\subsubsection{Cart localization}\label{sec:cart_loc}
After the cameras were synchronized and calibrated, the next step was to localize the cart in the scene. This was achieved by first identifying the cart on the static camera videos and then applying the camera matrices to obtain the metric coordinates. We are exploring other localization methods (e.g., visual odometry and ultra wide band positioning) and will continue to track progress on large-space localization. For the first batch of data included in our dataset, we manually labeled the locations of the cart.

\subsubsection{Pedestrian tracking and labeling}\label{sec:ped}
Similar to cart localization, we first tracked the pedestrians on the static camera videos and then identified their coordinates on the ground plane $G$. We found ByteTrack \cite{zhang2021bytetrack} to be very successful in tracking pedestrians in the image space. Upon human verification over our entire first batch of data, ByteTrack successfully aided the trajectory labeling of $91.8\%$ of the pedestrians automatically.

Once we obtained the automatically tracked labels in pixel space, we needed to convert them into metric space. With ByteTrack, each camera video contained a set of tracked trajectories in the image space $T_i=\{t_1,...,t_n\}, i\in\{1,2,3\}$ where $i$ is the camera index. We estimated the 3D trajectory coordinates for each pair of 2D trajectories $(t_i, t_j)| t_i\in T_i, t_j\in T_j, i\neq j$ and the set of estimated coordinates that resulted in the lowest reprojection error were selected to be the final trajectory coordinates in the metric space. 
%We then projected these 3D coordinates onto the ground plane $G$ to obtain the final metric coordinates.

Finally, we performed human verification over the entire tracking output, fixing any errors observed during the process. We also manually identified pedestrians that were outside our target tracking zone but had interactions with the pedestrians inside the tracking zone and included them as part of our dataset.

%Once we synchronize the time across all the captured videos, we generate labels on pedestrian positions and cart positions. We first perform a 3D reconstruction of the scene using Colmap \cite{schonberger2018robust} by additionally supplying it with dozens of static pictures of the atrium taken from a smartphone. During the 3D reconstruction process, Colmap also provides estimation of the cameras' poses. We can then estimate camera parameters by manually identify several 2D-3D point correspondences from the camera images and the 3D reconstruction. We use the method from \cite{lens-distort} to additionally estimate lens distortion coefficients. After all camera parameters are identified, we use SiamMask E \cite{chen2019siammaske} to perform tracking of the cart on videos captured by the three overlooking cameras. We manually check the tracking outcome and fix any errors observed. Once cart tracking is complete, we use triangulation to determine the 3D location of the cart. For pedestrians, we first run Detectron2 \cite{wu2019detectron2} to detect and track the pedestrians. Given the tracking results, we apply a set of heuristics based rules to fix tracking errors such as broken trajectories or double detection. We perform triangulation again to determine the 3D coordinates of the pedestrians given the tracked 2D coordinates captured by different cameras.
\section{Dataset Characteristics} \label{sec:evaluation}

\subsection{Comparison with Existing Datasets} \label{sec:eval-compare}

Compared to existing datasets collected in pedestrian natural environments, our TBD pedestrian dataset contains three components that greatly enhances the dataset's utility. These components are:

\textbf{Human verified labels grounded in metric space.} ETH \cite{ETH} and UCY \cite{UCY} datasets are often the only datasets to be included during the evaluation of various research models in many papers. This is largely because the trajectory labels in these datasets are human verified, unlike \cite{edinburgh}, \cite{cff}, \cite{grandcentral}, and \cite{atc} that solely rely on automatic tracking to produce labels. These trajectory labels are also grounded in metric space rather than image space (e.g. \cite{stanforddrone} and \cite{towncentre} only contain labels in bounding boxes). Having labels grounded in metric space eliminates the possibility that camera poses might have an effect on the scale of the labels. It also makes the dataset useful for robot navigation related research because robots plan in the metric space rather than image space.
    
\textbf{Combination of top-down views and perspective views.} Similar to datasets with top-down views, we use top-down views to obtain ground truth trajectory labels for every pedestrian present in the scene. Similar to datasets with perspective views, we gather perspective views from a ``robot" to imitate robot perception of human crowds. A dataset that contains both top-down views and perspective views will be useful for research projects that rely on perspective views. This allows perspective inputs to their models, while still having access to ground truth knowledge of the entire scene. 
%Examples include pedestrian motion prediction given partial observation of the scene and robot navigation research projects that only have onboard sensors as inputs to navigation models. 
    
\textbf{Naturalistic human behavior with the presence of a ``robot".} Unlike datasets such as \cite{lcas} or \cite{jrdb}, the ``robot" that provides perspective view data collection is a cart being pushed by human. As mentioned in section \ref{sec:hardware}, doing so reduces the novelty effects from the surrounding pedestrians. Having the ``robot" being pushed by humans also ensures safety for the pedestrians and its own motion has more natural human behavior.

\begin{table}[]
\caption{A survey of existing pedestrian datasets on how they incorporate the three components in section \ref{sec:eval-compare}. For component 1, a ``No" means either not human verified or not grounded in metric space. For component 2, TD stands for ``top-down view" and ``P" stands for ``perspective view".}
\label{tab:survey}
\begin{center}
\begin{tabular}{c||ccc}
\toprule
Datasets & Comp. 1 & Comp. 2 & Comp. 3\\
&   (metric labels) & (views) & (``robot") \\
\hline
TBD (Ours) & Yes & TD + P & Human + Cart \\
ETH \cite{ETH} & Yes & TD & N/A \\
UCY \cite{UCY} & Yes & TD & N/A \\
Edinburgh Forum \cite{edinburgh} & No & TD & N/A \\
VIRAT \cite{virat} & No & TD & N/A \\
Town Centre \cite{towncentre} & No & TD & N/A \\
Grand Central \cite{grandcentral} & No & TD & N/A \\
CFF \cite{cff} & No & TD & N/A \\
Stanford Drone \cite{stanforddrone} & No & TD & N/A \\
L-CAS \cite{lcas} & No* & P & Robot\\
WildTrack \cite{wildtrack} & Yes & TD & N/A\\
JackRabbot \cite{jrdb} & Yes & P & Robot\\
ATC \cite{atc} & No & TD & N/A\\
TH\"OR \cite{thor} & Yes & TD + P & Robot\\
\bottomrule
\end{tabular}
\end{center}
\vspace{-1.5em}
\end{table}

As shown in Table \ref{tab:survey}, current datasets only contain at most two of the three components\footnote{*L-CAS dataset does provide human verified labels grounded in the metric space. However, its pedestrian labels do not contain trajectory data, which means this dataset has limited usage in pedestrian behavior research.}. A close comparison is the TH\"OR dataset \cite{thor}, but its perspective view data are collected by a robot. Additionally, unlike all other datasets in Table \ref{tab:survey}, the TH\"OR dataset is collected in a controlled lab setting rather than in the wild. This injects artificial factors into human behavior, making them unnatural. 

\subsection{Dataset Statistics} \label{sec:eval-stats}

\begin{table}[ht]
\caption{Comparison of statistics between our dataset and other datasets that provide human verified labels grounded in the metric space. For total time length, 51 minutes of our dataset includes the perspective view data.}
\label{tab:stats}
\begin{center}
\begin{tabular}{c||ccc}
\toprule
Datasets & Time length & \# of pedestrians & Label freq (Hz)\\
\hline
\multirow{2}{*}{TBD (Ours)} & 133 mins & \multirow{2}{*}{1416} & \multirow{2}{*}{60} \\
     & (51 mins) & & \\
ETH \cite{ETH} & 25 mins & 650 & 15 \\
UCY \cite{UCY} & 16.5 mins & 786 & 2.5 \\
WildTrack \cite{wildtrack} & 200 sec & 313 & 2\\
JackRabbot \cite{jrdb} & 62 mins & 260 & 7.5\\
TH\"OR \cite{thor} & 60+ mins & 600+ & 100\\
\bottomrule
\end{tabular}
\end{center}
\end{table}

Table \ref{tab:stats} demonstrates the benefit of a semi-automatic labeling pipeline. With the aid of an autonomous tracker, humans only need to verify and make occasional corrections on the tracking outcomes instead of locating the pedestrians on every single frame. The data we have collected so far already surpassed all other datasets that provide human verified labels in the metric space in terms of total time, number of pedestrians and labeling frequency. We will continue this effort and collect more data for future works.

%It is worth noting that the effect of noise becomes larger with higher labeling frequency. We provide high frequency labeling so that more information and details can be available on the trajectories. When using our data, we recommend downsampling so that noise will have a lesser effect on pedestrian behavior modeling.

%As shown in Table 1, we have collected a significant amount of data using our current labelling pipeline (Section~\ref{sec:HDC_curr_pipeline}). We plan to to collect more data both at the current location and different locations as COVID-19 conditions relax. We will also incorporate  the ongoing improvements in Section~\ref{sec:HDC_new_pipeline} as  they are completed.

\subsection{Qualitative Pedestrian Behavior}

\begin{figure}[]
      \centering
      \includegraphics[scale=0.23]{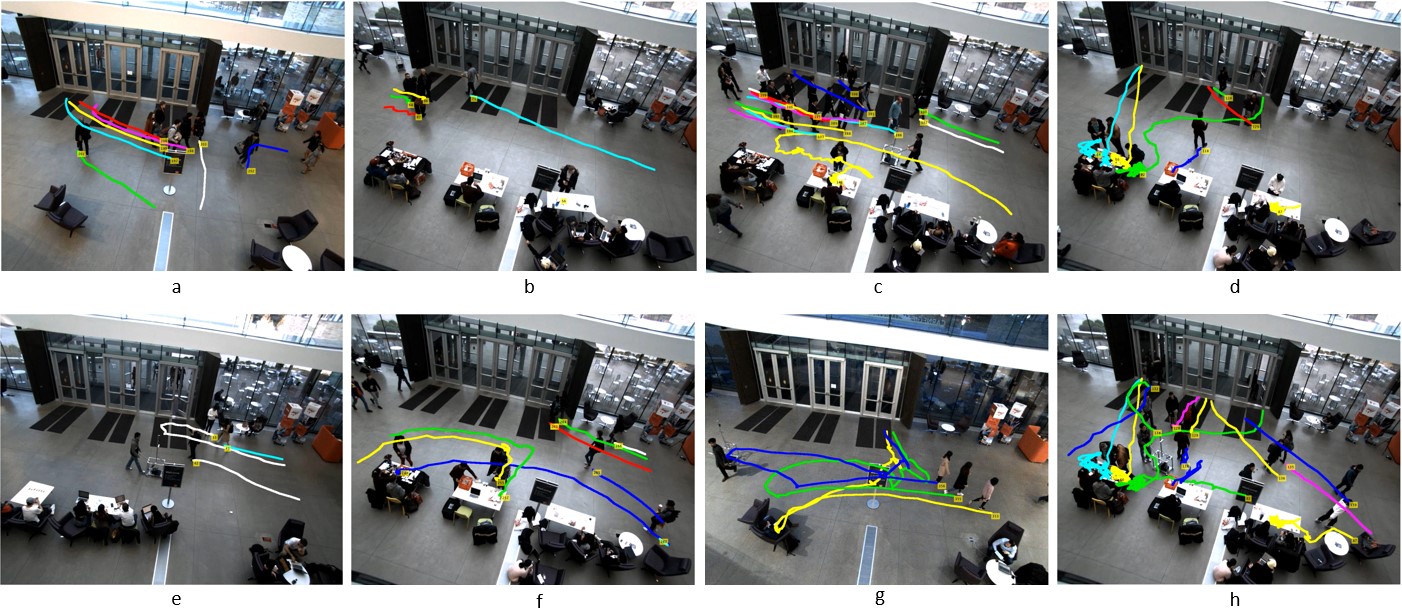}
      \caption{Example scenes from the TBD pedestrian dataset. a) a dynamic group. b) a static conversational group. c) a large tour group with 14 pedestrians. d) a pedestrian affecting other pedestrians' navigation plans by asking them to come to the table. e) pedestrians stop and look at their phones. f) two pedestrians change their navigation goals and turn towards the table. g) a group of pedestrians change their navigation goals multiple times. h) a crowded scene where pedestrians are heading towards different directions.}
      \label{fig:qual}
      \vspace{-1em}
   \end{figure}
   
Due to the nature of the environment where we collected the data, we observe a mixture of corridor and open space pedestrian behavior, many of which are rarely seen in other datasets. As shown in Figure \ref{fig:qual}, we observe both static conversation groups and dynamic walking groups. We also observe that some pedestrians naturally change goals mid-navigation. 

%Due to the timing of our data collection, we also observe ongoing activities where several students set up tables and engage people passing by. This activity produces additional interesting pedestrian interaction analogous to sellers touting and buyers browsing.
\section{Future Work}

%This paper presents a data collection system that is portable and enables large-scale data collection. Our systems offers better utility for pedestrian behavior research because our systems consists of human verified labels grounded in the metric space, a combination of both top-down views and perspective views, and a human-pushed cart that approximates naturalistic human motion with a socially-aware ``robot". We further couple the system setup with a semi-autonomous labeling process that easily produces human verified labels in order to meet the demands of the large-scale data collected by our hardware. Lastly, we present the TBD pedestrian dataset we have collected using our system, which not only surpasses the quantity of similar datasets, but also offers unique pedestrian interaction behavior that adds to the qualitative diversity of pedestrian interaction data.

A key concern about our current data collection setup is that our sensors consist purely of cameras. For better labeling accuracy, we are exploring adding a LiDAR to aid the autonomous tracking of pedestrians and adding an ultra wide band positioning system for better cart state estimation. We also plan to continue making improvements to our software system and underlying methods. Currently, the bottleneck to produce huge quantities of data still lies in correcting the few erroneous tracking outcomes of the automatic tracking procedures. A centralized user interface is under development to better document these tracking errors and to provide intuitive tools to fix them. As mentioned earlier, our approach enables additional data collection in a wide range of locations and constraints. Additional data collection and public updates to this initial dataset are planned.

\section*{ACKNOWLEDGMENT}
This work was supported by grants (IIS-1734361 and IIS-1900821) from the National Science Foundation.

%%%%%%%%%%%%%%%%%%%%%%%%%%%%%%%%%%%%%%%%%%%%%%%%%%%%%%%%%%%%%%%%%%%%%%%%%%%%%%%%
{
\bibliographystyle{IEEEtranS}
\bibliography{IEEEabrv}
}

\end{document}